\documentclass{article}

\PassOptionsToPackage{sort}{natbib}
\usepackage{iclr2027_conference,times}
\usepackage[final]{microtype}
\usepackage[hidelinks]{hyperref}
\usepackage{booktabs}
\usepackage{graphicx}
\usepackage{amsmath,amssymb}
\usepackage{wrapfig}

\newcommand{\method}{\textsc{SPARC}}
\newcommand{\x}{\mathbf{x}}
\newcommand{\y}{\mathbf{y}}
\newcommand{\sg}{\operatorname{sg}}

\title{Transferable Mass Spectrum Prediction via Reference-Guided Test-time Specialization}

\author{%
Yunhua Zhong$^{1,2,\dagger}$,
Runting Li$^{1,3}$,
Yifan Li$^{1}$,
Pan Liu$^{1}$,\\
\textbf{Zhiwen Yang$^{1,4}$, 
Zikun Wang$^{1}$,
Yixuan Tang$^{1,5}$,
Jun Xia$^{1,6,*}$} \\
$^{\dagger}$This work was done during an undergraduate internship at HKUST-GZ.\\
$^{1}$The Hong Kong University of Science and Technology (Guangzhou)\\
$^{2}$The University of Hong Kong,
$^{3}$South China University of Technology\\
$^{4}$Hong Kong Polytechnic University,
$^{5}$Jinan University\\
$^{6}$The Hong Kong University of Science and Technology\\
$^{*}$Corresponding Author.\\
}

\iclrfinalcopy

\begin{document}
\maketitle

\begin{abstract}
    Tandem mass spectrum prediction supports compound identification across metabolomics, natural-product discovery, and environmental analysis. However, pretrained predictors often degrade under shifts in chemical space and acquisition conditions, while retraining domain-specific models from scratch is costly. We introduce SPARC, a retrieval-guided test-time specialization framework that adapts a pretrained predictor using a spectral reference library without accessing test-query spectra. For each target query, SPARC retrieves chemically related reference spectra to recalibrate fragment intensities within the learned fragmentation space. During Transfer, SPARC combines reference-guided spectral adaptation with reliability-aware consistency, using reconstruction behavior on retrieved spectra to selectively preserve trustworthy predictions during continual specialization. Across MassSpecGym, NPLIB1 and application-specific GNPS libraries, SPARC improves spectral prediction under multiple transfer settings. These results establish retrieval-guided test-time specialization as a practical strategy for extending pretrained MS/MS predictors to specific chemical and acquisition domains, with continual test-time training providing further refinement during deployment.
\end{abstract}

\section{Introduction}

Tandem mass spectrometry is central to molecular annotation, natural-product discovery \citep{wang2016gnps,duhrkop2019sirius4} and environmental analysis \citep{schymanski2014confidence}. By estimating fragment-ion masses and intensities from molecular structures, molecule-to-spectrum predictors extend spectral reference coverage beyond available experimental measurements \citep{murphy2023graffms,young2024massformer,iceberg_repo,nowatzky2025fiora}.
Their performance, however, depends on the chemical and acquisition distributions represented during training \citep{bremer2022cfmid, young2024massformer}. Large spectral libraries expose a model to broader chemical space and fragmentation behavior than a single study can provide \citep{murphy2023graffms, gupta2026spectraverse}. Real-world deployment rarely queries this space uniformly. Each application emphasizes a particular molecular distribution together with its own adducts, ionization modes, instruments, collision energies, and metadata. These factors can change spectral similarity and prediction performance \citep{bremer2022cfmid,hoang2024platforms, liu2025adduct}. Consequently, broad training coverage does not ensure accurate predictions for the chemical space and acquisition condition of a particular study.

This mismatch produces a practical generalization problem. Changes in acquisition conditions alter the mapping from molecular structure to fragment intensities \citep{bremer2022cfmid}. Study-defined molecular distributions can also concentrate predictions in regions that were sparse during source training \citep{bushuiev2024massspecgym}. Adduct-dependent fragmentation provides a concrete example. Compounds retained under one precursor ion may lack paired spectra under another condition. The same molecule can also fragment differently across adducts \citep{schmid2021iimn}, creating coverage and chemical-composition gaps. When measured spectra are unavailable for the target molecules, chemically related reference spectra provide an alternative source of domain-specific information. Prior work on spectral representation learning and analogue search has shown that MS/MS neighborhoods can recover chemically related molecules \citep{huber2021spec2vec,dejonge2023ms2query}. This motivates test-time specialization that adapts the predictor to the queried domain.

\begin{figure*}[!htbp]
  \centering
  \includegraphics[width=\textwidth]{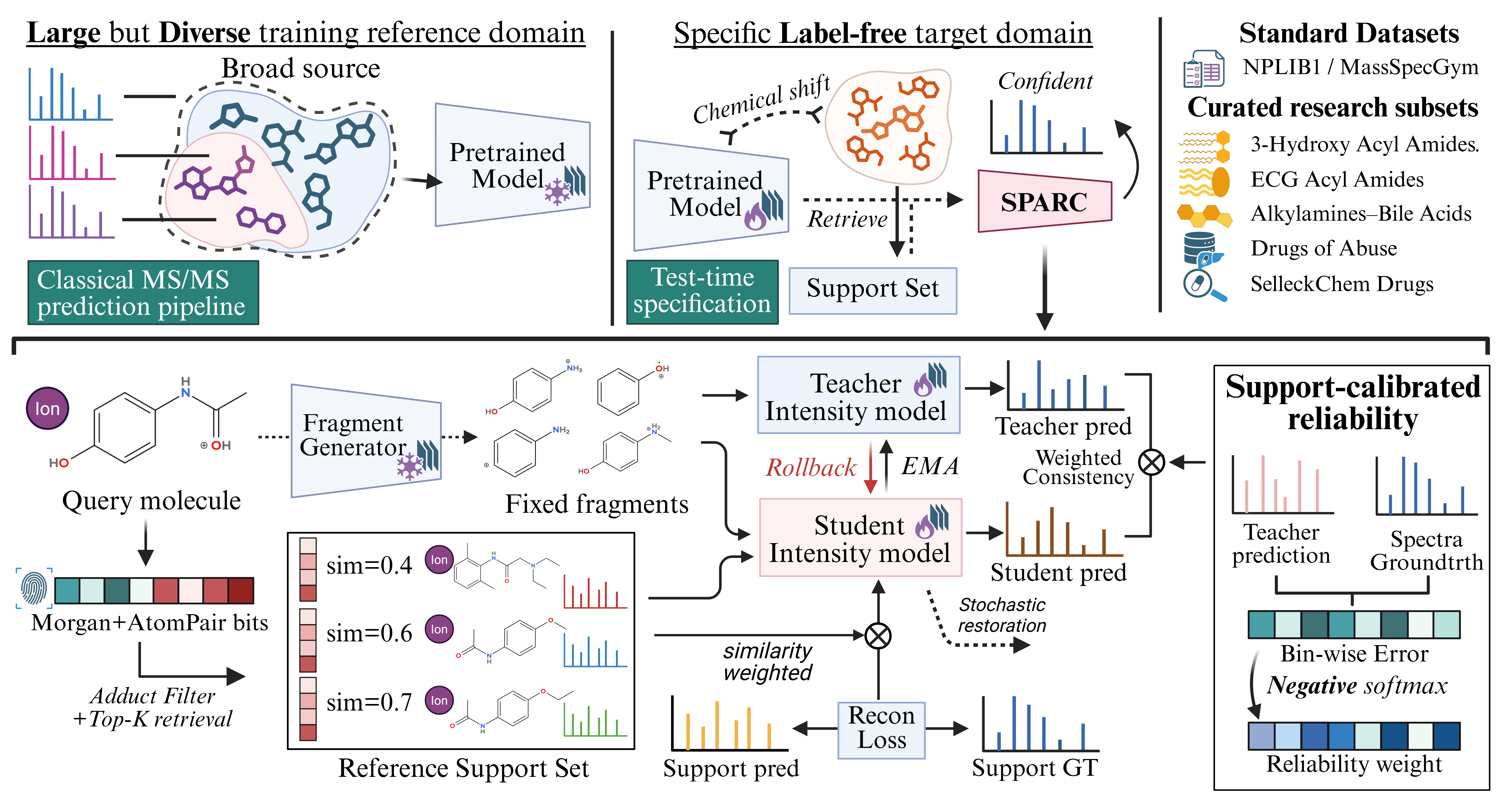}
  \caption{Overview of \method. \textbf{Top:} A predictor pretrained on a broad source collection is specialized to a target research domain. \textbf{Bottom:} Target molecules retrieve acquisition-compatible and chemically similar support spectra. The fragment generator remains frozen, while support reconstruction errors calibrate bin-wise reliability of an exponential moving average teacher. The student intensity model optimizes similarity-weighted support reconstruction and reliability-weighted teacher--student consistency, with stochastic restoration and rollback constraining adaptation drift.}
  \label{fig:compass-overview}
\end{figure*}

Test-time training specializes a pretrained model during inference. In classification, TENT usesprediction entropy as a confidence signal, while CoTTA combines teacher averaging with stochastic restoration to limit error accumulation and forgetting \citep{wang2021tent,wang2022cotta}. Transferring these ideas to MS/MS prediction, however, requires complementing confidence-based adaptation with an external spectral signal indicating how predictions change.

Spectral entropy summarizes the dispersion of normalized peak intensities \citep{li2021spectralentropy}, but it does not encode whether peaks occur at the correct locations. Entropy can therefore decrease by suppressing weak peaks without improving the predicted fragmentation pattern. Moreover, target molecular structures identify the chemical region being queried but do not specify how the predicted spectrum should change. Chemically related measured reference spectra provide the missing external signal, while teacher predictions regularize the update and preserve reliable source-model structure.

We introduce SPARC (\textbf{S}pectral \textbf{P}rediction via \textbf{A}daptation with \textbf{R}etrieval-calibrated \textbf{C}onsistency), a retrieval-guided test-time training framework that combines support-guided specialization with drift-aware continual adaptation for pretrained MS/MS predictors (Fig. \ref{fig:compass-overview}). For each target, \method \; retrieves chemically related labeled spectra using acquisition-aware prioritization and uses their measured intensities to supervise adaptation. In addition, \method{} uses reconstruction errors on retrieved support spectra to weight teacher--student consistency, relaxing teacher constraints in bins with larger support errors. An exponential moving average teacher, stochastic restoration and rollback mechanism constrain continual updates together.

We evaluate SPARC on MassSpecGym and NPLIB1 \citep{duhrkop2021canopus, bushuiev2024massspecgym}, covering specialization to an adduct-defined subset and transfer between spectral libraries. We additionally curate five application-specific GNPS collections using a shared processing pipeline to represent focused scientific domains \citep{wang2016gnps}. These evaluations show robust improvements in the main benchmark transfer settings and higher entropy similarity across all application libraries. Peak-level analyses further characterize how adaptation improves predictions within the preserved candidate space, revealing increases in precision and F1-score at the peak and intensity level.

\section{Related Work}
\paragraph{MS/MS Spectrum Prediction.}
Computational mass spectrometry encompasses both spectrum interpretation and molecule-to-spectrum prediction. SIRIUS combines isotope patterns and fragmentation trees for molecular formula and structure annotation \citep{duhrkop2019sirius4}. Earlier deep learning approaches focused on direct spectrum regression, including NEIMS for fingerprint-based EI prediction \citep{wei2019rapid}, 3DMolMS for 3D structure-based MS/MS prediction \citep{hong2023dmolms}, and GrAFF-MS for graph-based high-resolution spectrum prediction \citep{murphy2023graffms}. More recently, explicit fragmentation models have further connected predicted peaks to molecular substructures. Iceberg combines autoregressive fragmentation-graph generation with fragment intensity prediction \citep{iceberg_repo}, whereas FIORA estimates fragment-ion probabilities from single-bond cleavages and their local molecular neighborhoods \citep{nowatzky2025fiora}. MassFormer uses graph transformers to model molecular structure, whereas FraGNNet introduces a structured probabilistic model for high-resolution spectrum prediction \citep{young2024massformer, young2024fragnnet}.

\paragraph{Test-Time Training.}
Test-time training updates a trained model using information available during inference. TENT minimizes prediction entropy by updating normalization parameters \citep{wang2021tent}, while CoTTA supports non-stationary streams through an exponential-moving-average teacher, augmentation pseudo-labels, and stochastic restoration \citep{wang2022cotta}. Beyond computer vision, TAIP adapts interatomic potentials to out-of-distribution molecular configurations using global- and local-structure self-supervision \citep{cui2025taip}. Test-time learning has also been explored in mass spectrometry for peptide-spectrum prediction \citep{ye2024testtime} and \emph{de novo} small-molecule generation from observed spectra \citep{mismetti2026testtime}. 

\paragraph{Mass Spectral Libraries and Benchmarks.}
Public repositories such as GNPS, MassBank, and MetaboLights aggregate spectra across laboratories, scientific domains, and acquisition conditions, providing broad coverage but substantial heterogeneity \citep{wang2016gnps, horai2010massbank, yurekten2024metabolights}. To support systematic machine learning evaluation, several curated benchmarks have since been developed. NPLIB1 provides a natural-product-oriented collection, MassSpecGym standardizes prediction and retrieval benchmarks, and MS$^n$Lib and SpectraVerse broaden coverage across compounds, adducts, and ionization modes \citep{duhrkop2021canopus, bushuiev2024massspecgym, brungs2025msnlib, gupta2026spectraverse}. In addition, application-specific collections such as oxylipin libraries and GNPS community subsets reflect concrete scientific settings \citep{elloumi2024neomsms}, from which we curate five application-grounded target domains for SPARC.

\section{Method}
\label{method:overall}

We consider MS/MS spectrum prediction from a known molecular structure and its acquisition conditions. Let $\x=(\mathcal{G},\mathbf{c})$, where $\mathcal{G}$ is the molecular graph and $\mathbf{c}$ contains the available metadata, including precursor adduct, instrument type, and collision energy. The output is a non-negative intensity vector $\y\in\mathbb{R}_{\geq 0}^{B}$ over $B$ fixed $m/z$ resolution bins. 

A predictor $F_\omega$ estimates the intensity vector as $\hat{\y}=F_{\omega}(\x)$. At deployment, the target molecules and acquisition conditions may differ from those represented during source training. Given a pretrained predictor $F_{\omega_0}$, a target query $x_t$, and a labeled reference library $S=\{(x_i^s,y_i^s)\}_{i=1}^{M}$, our goal is to specialize the predictor using the query input and reference spectra without using the measured query spectrum in the adaptation objective.

We consider two deployment settings, both with access to a labeled reference library. In the \textbf{target-with-validation} setting, labeled target-domain validation spectra guide update acceptance and checkpoint selection. In the \textbf{target-only} setting, no target validation labels are used, and updates are accepted by an entropy-based guard. Measured test-query spectra are reserved for evaluation only.

\subsection{Fragment-Space Preserving Adaptation}
\label{sec:fixed-fragment-space}

\method \; instantiates $F_{\omega}$ with Iceberg, which separates discrete fragment generation from continuous intensity prediction \citep{iceberg_repo}. Writing $\omega=(\psi,\theta)$ for the parameters of these two stages respectively, we have
\begin{equation}
  \mathcal{H}_{\x}=G_{\psi}(\x),
  \qquad
  \hat{\y}=f_{\theta}(\x,\mathcal{H}_{\x}).
  \label{eq:iceberg-factorization}
\end{equation}
Here, $G_{\psi}$ autoregressively constructs a directed acyclic graph fragmentation $\mathcal{H}_{\x}$ of candidate fragments, and $f_{\theta}$ predicts their contributions to the spectrum. The source checkpoint provides the initial parameters $(\psi_0,\theta_0)$. Iceberg-Generate supplies a ranked set of candidate fragments to Iceberg-Score, allowing us to reuse the pretrained generator as a fixed fragmentation prior. Accordingly, \method{} fixes $G_{\psi_0}$ and adapts only $\theta$. Conditioned on molecular and fragment representations and acquisition metadata, the intensity model uses chemically related reference spectra to learn target spectral patterns, emphasizing characteristic fragments and suppressing less relevant candidates. Adaptation therefore changes intensity allocation within the existing candidate space.

The adaptation pipeline uses retrieved spectra in two complementary ways. Their measured intensities directly supervise the student, while the reconstruction errors of an EMA teacher determine the bin-wise weights for consistency on the queries. Student and teacher states persist across successive queries, with stochastic restoration and rollback governing continual updates.

\subsection{Condition-Aware Chemical Retrieval}
\label{sec:chemical-retrieval}

\method \; constructs the chemical-local support neighborhood for each query, so the supervision follows the chemical region being processed. For a molecule $\x$, let $\phi(\x)$ be the $\ell_2$-normalized concatenation of its Morgan and hashed AtomPair bit fingerprints \citep{rogers2010ecfp,carhart1985atompairs}. Each reference spectrum is scored by its fingerprint cosine similarity to the target molecule.
\begin{equation}
r_{t,i} = \frac{\phi(x_t)^\top \phi(x_i^s)}{ \|\phi(x_t)\|_2\|\phi(x_i^s)\|_2}
  \label{eq:retrieval-score}
\end{equation}

Acquisition compatibility takes precedence over chemical ranking. With the default adduct-based retrieval, references matching the query adduct form the candidate pool. Within this pool, references also matching the query instrument are ranked first, followed by the remaining adduct-compatible references. Each group is sorted by $r_{t,i}$, and the first $K$ entries form $S_t$. The implementation falls back to global retrieval only when the adduct-compatible pool is empty. Collision energy and precursor $m/z$ remain model inputs but are not hard retrieval filters.

Retrieved spectra contribute according to their chemical relevance. Writing $K_t=|\mathcal{S}_t|$ and reindexing locally reindexing the entries of $\mathcal{S}_t$, we define
\begin{equation}
  a_{t,i}=\frac{\exp(r_{t,i}/\tau)}
  {\sum_{j=1}^{K_t}\exp(r_{t,j}/\tau)},
  \qquad \tau>0,
  \label{eq:support-weights}
\end{equation}
and minimize the similarity-weighted support reconstruction loss
\begin{equation}
  \mathcal{L}_{\mathrm{sup}}(\theta;\mathcal{S}_t)=
  \frac{1}{B}\sum_{i=1}^{K_t}a_{t,i}
  \left\|f_{\theta}(\x_i^s)-\y_i^s\right\|_2^2.
  \label{eq:support}
\end{equation}
This term anchors adaptation in measured spectra rather than relying exclusively on the model’s own predictions.

\subsection{Support-Calibrated Teacher Consistency}
\label{sec:support-calibration}

The teacher supplies a target prediction, but its reliability need not be
uniform across the spectrum \citep{tarvainen2017meanteacher}. We initialize the student $\theta$ and teacher
$\bar\theta$ from $\theta_0$. Before each update, we evaluate the teacher on the retrieved labeled spectra and estimate an error profile across $m/z$ bins:
\begin{equation}
  \begin{aligned}
    e_{t,b}=\frac{1}{K_t}\sum_{i=1}^{K_t}
    [f_{\bar\theta}(\x_i^s)_b-y_{i,b}^s]^2, \qquad c_{t,b}=\exp(-\gamma e_{t,b}), \gamma\geq 0.
  \end{aligned}
  \label{eq:bin-confidence}
\end{equation}
Support errors are averaged uniformly over the retrieved set, whereas retrieval-similarity weights modulate the supervised reconstruction loss. Chemical locality enters the error estimate through support selection. The resulting $c_{t,b}$ is computed without gradients and controls the strength of teacher consistency on the shared $m/z$ grid: larger support reconstruction errors yield weaker constraints. We use these weights as a support-derived regularization signal, rather than as calibrated probabilities of correctness for individual query fragments.

We regularize the student toward the unperturbed teacher prediction,
reducing pressure in bins that the teacher reconstructs poorly:
\begin{equation}
\mathcal{L}_{\mathrm{con}} = \frac{1}{B}
\sum_{b=1}^{B}c_{t,b}\left[f_{\theta}(x_t)_b-\operatorname{sg}\!\left(f_{\bar{\theta}}(x_t)_b\right)\right]^2
  \label{eq:consistency}
\end{equation}
where $\sg$ denotes stop-gradient. The core adaptation objective is
\begin{equation}
  \mathcal{L}_{\mathrm{adapt}}=
  \mathcal{L}_{\mathrm{con}}+
  \lambda_{\mathrm{sup}}\mathcal{L}_{\mathrm{sup}}
  \label{eq:adaptation-objective}
\end{equation}
where $\lambda_{sup}$ controls the weight of direct support supervision. The measured support spectra drive student updates, while teacher consistency limits changes on the current query according to  reliability. These weights determine how student model is constrained by the teacher at each bin. 


\subsection{Continual Updates and Drift Control}
\label{sec:continual-updates}

\paragraph{Mechanisms for stabilizing continual updates.} \method\;uses two CoTTA-inspired mechanisms to stabilize continual updates, an EMA teacher and stochastic restoration \citep{wang2022cotta}. For each query and its retrieved support set, the student takes gradient steps on Eq.~\eqref{eq:adaptation-objective}, and the accepted state is carried to the next query. At inner step $u$, the optimizer first produces a trial student state $\theta_u^{+}$; the teacher then tracks this
state according to
\begin{equation}
  \bar\theta_{u+1}=\alpha\bar\theta_u+(1-\alpha)\theta_u^{+}.
  \label{eq:teacher-update}
\end{equation}
The EMA decay coefficient $\alpha$ is fixed in our implementation. After the teacher update, each
trainable student weight or bias entry is independently restored to its source
value with probability $\rho$:
\begin{equation}
  \begin{aligned}
    m_{u,j}\sim\operatorname{Bernoulli}(\rho), \qquad \qquad\theta_{u+1,j}=m_{u,j}\theta_{0,j}
    +(1-m_{u,j})\theta_{u,j}^{+}.
  \end{aligned}
  \label{eq:stochastic-restoration}
\end{equation}
Thus, the teacher averages the optimized student before restoration, whereas restoration anchors the student to the pretrained parameters. This ordering smooths the teacher trajectory while limiting cumulative drift in the model that is updated online.

\paragraph{Supervised and unsupervised rollback.} Besides stabilizer, \method\;controls the remaining drift with a rollback rule whose evidence depends on
whether labeled validation spectra are available. Before each trial update, we snapshot the whole model and optimizer states. 

When validation spectra are available, a trial update is accepted only if its
post-update validation CosSim remains at least $\eta$ times the pre-update value. This comparison uses validation spectra for the rollback decision, while the adaptation loss remains unchanged.

When validation spectra are unavailable, the same rollback operation uses a label-free entropy guard instead. For a predicted intensity vector $\mathbf{z}$, define normalized spectral entropy as
\begin{equation}
  \begin{aligned}
    p_b(\mathbf{z})&=\frac{\max(z_b,0)}
    {\sum_{j=1}^{B}\max(z_j,0)+\epsilon},\\
    h(\mathbf{z})&=-\frac{1}{\log B}\sum_{b=1}^{B}
    p_b(\mathbf{z})\log[p_b(\mathbf{z})+\epsilon].
  \end{aligned}
  \label{eq:guard-entropy}
\end{equation}

Let $\bar{y}^{-}_{u}$ and $\bar{y}^{+}_{u}$ denote the unperturbed teacher predictions for the current query before and after the trial update,
respectively. We define
\begin{equation}
H^{-}_{u}=h(\bar{y}^{-}_{u}), \qquad H^{+}_{u}=h(\bar{y}^{+}_{u}).
  \label{eq:batch-guard-entropy}
\end{equation}
We reject the update when $H^{+}_{u}-H^{-}_{u}>\delta$. This label-free guard rejects updates that abruptly broaden the predicted intensity distribution, providing a lightweight stability criterion.

\section{Experiments}

\paragraph{Datasets and benchmarks}
\label{experiment:benchmark}
Following prior work, we first evaluate \method{} on MassSpecGym (MSG) \citep{bushuiev2024massspecgym} and NPLIB1 \citep{duhrkop2021canopus}. We consider adduct-focused specialization from the full MSG dataset to MSG-MNa subset and cross-library transfer from MSG to NPLIB1. MSG$\rightarrow$MSG-MNa is an adduct-defined subset specialization, as the full-MSG source checkpoint includes M+Na examples. We further use five experimental libraries on GNPS: Drugs of Abuse, 3-Hydroxy Acyl Amides, ECG Acyl Amides C4--C24, Alkylamines--Bile Acids, and SelleckChem FDA. These collections span forensic toxicology, acyl-amide and lipid chemistry, bile-acid conjugates, and approved pharmaceuticals. For all five GNPS target libraries, the MSG training split serves as the labeled reference support library, while target spectra are reserved only for evaluation. To prevent information leakage, Murcko scaffolds were computed for molecules, and support molecules sharing a scaffold with any validation or test molecule were excluded before retrieval \citep{bemis1996murcko}. Complete preprocessing setting are provided in Supplementary. 

\paragraph{Evaluation}
We evaluate spectral fidelity using cosine similarity (CosSim), Jensen--Shannon entropy similarity (EntSim), and mean squared error (MSE), with entropy as an auxiliary diagnostic. Before scoring, we retain the 100 highest-intensity predicted bins. Comparisons use the successful intersection within methods, and coverage is reported in Supplementary. We additionally evaluate peak-presence and ion-current quality using precision, recall, and F1 after exact-bin matching, with peak- and intensity-level metrics computed from matched prediction and ground-truth. 

\paragraph{Implementation details and baselines}
\method \; retrieves 64 support spectra for each query. During adaptation, \method\;optimizes its objective with Adam using a learning rate of \(1 \times 10^{-4}\). Each molecular representation concatenates 2,048 radius-2 Morgan fingerprint bits with 3,072 hashed atom-pair bits. We set the support-loss coefficient to \(\lambda_{\mathrm{sup}}=1\) and use \(\tau=0.1\), \(\gamma=10\), \(\alpha=0.99\), \(\rho=0.02\), and \(\delta=0.1\) for the remaining adaptation and regularization terms. Finally, \method\; use student model for prediction. We implement all models on NVIDIA A800 (80GB) GPUs.

We compare against NEIMS, GrAFF-MS, FIORA and pretrained Iceberg models as predictor baselines, and TENT and CoTTA on Iceberg as source-free adaptation baselines. Architectural details are reported in Supplementary. 

\paragraph{Adaptation protocols.}
For the MSG and NPLIB1 benchmarks, labeled target-training spectra provide
support supervision. \method{}(val) specializes the source checkpoint
using target-validation spectra for rollback and checkpoint selection,
and is then evaluated without test-time updates. \method{}--TTT starts
from the same selected checkpoint and continues adapting on test-query
structures and acquisition metadata using entropy-based rollback.
The GNPS experiments instead omit validation specialization and use
MSG-training spectra as the reference library, with entropy-based rollback
throughout online adaptation. Measured test-query spectra are never used
in adaptation losses, rollback decisions, or checkpoint selection.

\section{Results}
\subsection{Prediction heterogeneity and spectral domain shifts.}


\paragraph{Cross-adduct shift changes both chemistry and fragmentation.} In MassSpecGym, only 19.0\% of molecules with M+H spectra also have M+Na spectra. These molecules are enriched for N-free, polyol/carbohydrate-like, and O-rich structures, but depleted for basic amines, aromatic nitrogen, and sulfone/sulfonamides, consistent with known adduct preferences \citep{kruve2013sodium}. Even among matched molecules, over 50\% share no peaks and few neutral losses, reflecting strong adduct-dependent fragmentation \citep{liu2025adduct, dejonge2026crossionization}. M+Na spectra also show lower entropy. Detailed analysis are provided in Supplementary. 

\paragraph{Locality motivates retrieval.}

We next ask whether the prediction error is uniform. We visualize this by projecting DreaMS embeddings and molecule fingerprints of dataset samples into a two-dimensional UMAP and coloring each sample by its prediction performance \citep{bushuiev2026self}. Regions of high and low accuracy occur within every split and in both molecule and spectra views (Fig.~\ref{fig:performance-locality}). This qualitative pattern is supported in Supplementary. Because errors cluster locally, fine-tuning on the whole support set can improve performance across the entire local neighborhood, turning test-time training into genuine domain specialization \citep{huber2021ms2deepscore}. 

\begin{figure}[t]
  \centering
  \includegraphics[width=0.9\columnwidth]{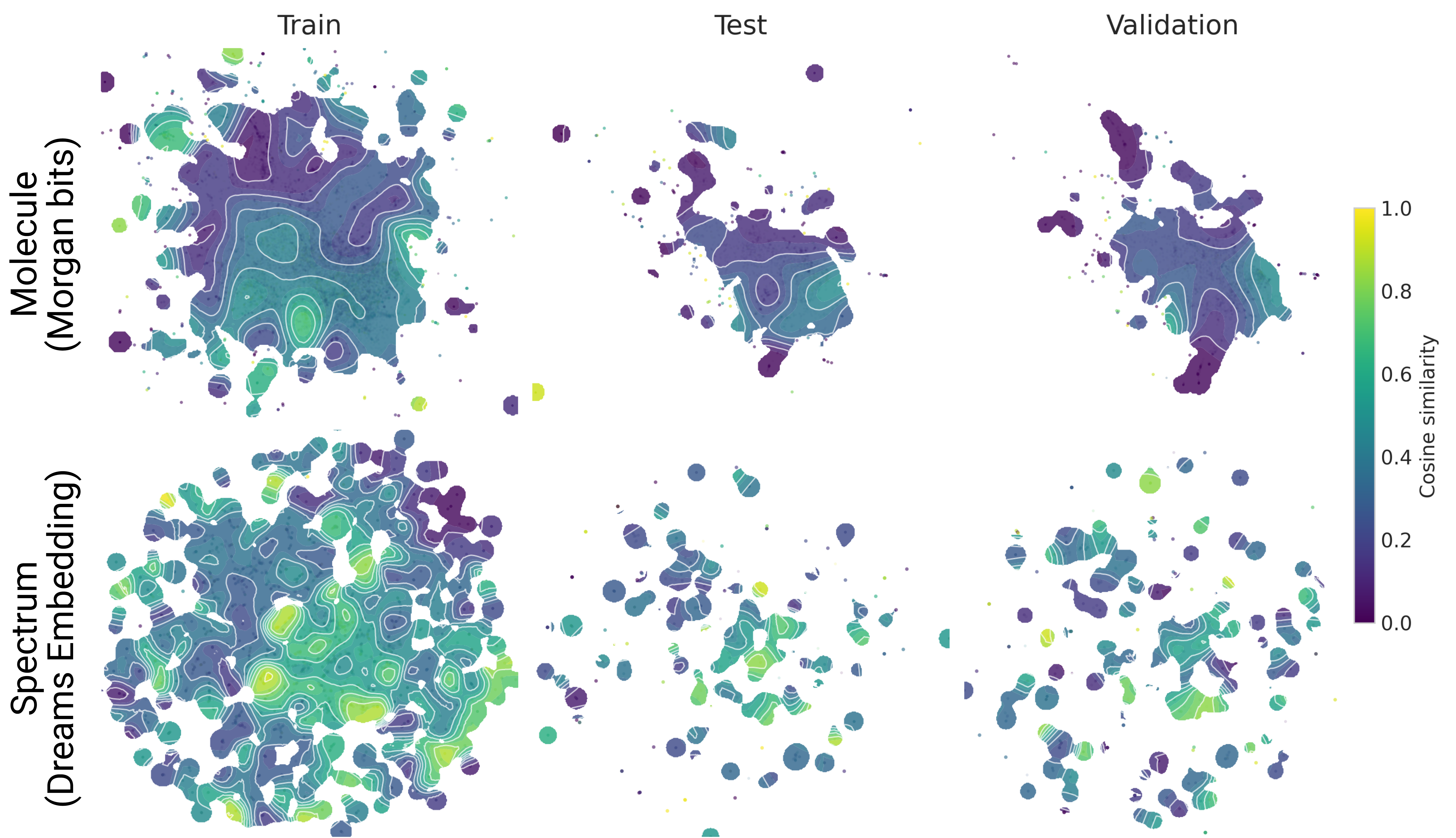}
  \caption{\textbf{Prediction performance is locally structured in spectral and molecule spaces.}}
  \label{fig:performance-locality}
\end{figure}

\subsection{\method\;Specializes across and within Domains}

We first evaluate SPARC under explicit domain shifts. On MSG$\rightarrow$MSG-MNa, SPARC raises EntSim and CosSim from 0.180/0.280 for the pretrained MSG checkpoint to 0.274/0.370 after validation-stage specialization, with \method--TTT further reaching a CosSim of 0.376 (Table~\ref{tab:transfer}). The adapted model also exceeds the Iceberg model trained directly on M+Na, which obtains 0.262/0.348. SPARC therefore recovers a substantial part of the adduct-specific prediction gap through reference-guided specialization of the pretrained model.

\begin{table}[!hbtp]
\centering
\caption{Test performance for MSG$\rightarrow$MSG-MNa and MSG$\rightarrow$NPLIB1 transfer.}
\label{tab:transfer}
\scriptsize
\setlength{\tabcolsep}{2pt}
\renewcommand{\arraystretch}{0.9}
\resizebox{\columnwidth}{!}{%
\begin{tabular}{@{}c@{\hspace{4pt}}c@{}}
\begin{tabular}[t]{lrrrr}
    \toprule
    Method & Entropy & MSE & EntSim & CosSim \\
    \midrule
    \multicolumn{5}{c}{\textit{MSG$\rightarrow$MSG-MNa (MSG initialized)}} \\
    \midrule
    NEIMS          & 4.150 & 14.461 & 0.072 & 0.072 \\
    GrAFF-MS       & 4.134 & 15.822 & 0.075 & 0.071 \\
    FIORA          & 3.015 & 1.020  & 0.063 & 0.071 \\
    Iceberg--MSG   & 3.461 & 1.242 & 0.180 & 0.280 \\
    Iceberg--MNa   & 2.485 & 1.006 & 0.262 & 0.348 \\
    \method(val)   & 2.715 & \underline{0.895} & \underline{0.274} & \underline{0.370} \\
    \method-TTT    & 2.923 & \textbf{0.882} & \textbf{0.284} & \textbf{0.376} \\
    \midrule
    Iceberg(tent)  & 2.077 & 1.183 & 0.229 & 0.290 \\
    Iceberg(cotta) & 3.419 & 1.189 & 0.191 & 0.288 \\
    \bottomrule
\end{tabular}
&
\begin{tabular}[t]{lrrrr}
    \toprule
    Method & Entropy & MSE & EntSim & CosSim \\
    \midrule
    \multicolumn{5}{c}{\textit{MSG$\rightarrow$NPLIB1 (MSG initialized)}} \\
    \midrule
    NEIMS            & 3.568 & 20.246 & 0.392 & 0.363 \\
    GrAFF-MS         & 3.396 & 21.170 & 0.330 & 0.280 \\
    FIORA            & 1.900 & 1.317 & 0.369 & 0.412 \\
    Iceberg-MSG      & 3.586 & 1.269 & 0.470 & 0.517 \\
    Iceberg-NPLIB1   & 3.516 & 1.217 & 0.521 & 0.586 \\
    \method(val)     & 3.112 & \underline{1.063} & \underline{0.559} & \underline{0.630} \\
    \method--TTT     & 3.128 & \textbf{1.041} & \textbf{0.575} & \textbf{0.650} \\
    \midrule
    Iceberg(tent)    & 3.052 & 1.252 & 0.490 & 0.517 \\
    Iceberg(cotta)   & 3.560 & 1.245 & 0.472 & 0.518 \\
    \bottomrule
\end{tabular}
\end{tabular}%
}
\end{table}

Starting from the MSG checkpoint, specialization using NPLIB1 training spectra as support and validation-guided rollback and checkpoint selection improves EntSim and CosSim from 0.470/0.517 to 0.559/0.630 (Table \ref{tab:transfer}). Across both transfer settings, validation-guided specialization provides the main performance gain, with test-stream continuation yielding additional refinement as target queries arrive. Thus, the benefit is not confined to the adduct subset but persists when transferring to a distinct library. A separate controlled comparison matches source initialization, target-training support access, validation protocol, and held-out test evaluation to isolate the benefit beyond retrieval-only fine-tuning. On NPLIB1, similarity-weighted retrieved-support fine-tuning achieves 0.5873 CosSim, compared with 0.5545 for global fine-tuning, supporting the value of chemically conditioned training beyond access to labeled target-domain spectra alone. 

\begin{table}[!htbp]
\centering
\caption{Test performance for MSG M+H and NPLIB1.}
\label{tab:in_domain}
\scriptsize
\setlength{\tabcolsep}{2pt}
\renewcommand{\arraystretch}{0.9}
\resizebox{\columnwidth}{!}{%
\begin{tabular}{@{}c@{\hspace{4pt}}c@{}}
\begin{tabular}[t]{lrrrr}
    \toprule
    Method & Entropy & MSE & EntSim & CosSim \\
    \midrule
    \multicolumn{5}{c}{\textit{MSG M+H (MSG initialized)}} \\
    \midrule
    NEIMS         & 3.982 & 16.180 & 0.184 & 0.162 \\
    GrAFF-MS      & 3.756 & 16.713 & 0.169 & 0.137 \\
    FIORA         & 1.742 & 0.917  & 0.434 & 0.460 \\
    Iceberg--MSG  & 3.477 & \textbf{0.853} & 0.462 & 0.529 \\
    \method(val)  & 2.866 & 0.907 & \underline{0.494} & \underline{0.532} \\
    \method--TTT  & 2.794 & \underline{0.880} & \textbf{0.505} & \textbf{0.547} \\
    \bottomrule
\end{tabular}
&
\begin{tabular}[t]{lrrrr}
    \toprule
    Method & Entropy & MSE & EntSim & CosSim \\
    \midrule
    \multicolumn{5}{c}{\textit{NPLIB1 (NPLIB1 initialized)}} \\
    \midrule
    NEIMS          & 3.349 & 20.652 & 0.324 & 0.280 \\
    GrAFF-MS       & 3.521 & 19.689 & 0.384 & 0.362 \\
    FIORA          & 1.596 & 1.237 & 0.419 & 0.484 \\
    Iceberg-NPLIB1 & 3.516 & 1.217 & 0.521 & 0.586 \\
    \method(val)   & 2.925 & \underline{1.107} & \underline{0.559} & \underline{0.616} \\
    \method--TTT   & 2.944 & \textbf{1.092} & \textbf{0.565} & \textbf{0.628} \\
    \bottomrule
\end{tabular}
\end{tabular}%
}
\end{table}

SPARC can also refine predictors that are already matched. We apply the same adaptation procedure within MSG M+H and within NPLIB1, where the backbone was trained on the target domain itself (Table~\ref{tab:in_domain}). On the dominant MSG M+H subset, \method{}\;increases EntSim and CosSim from 0.462/0.529 to 0.505/0.547, and increases EntSim and CosSim from 0.521/0.586 to 0.565/0.628 on NPLIB1. These within-domain gains show that retrieved reference spectra provide a useful local refinement signal even after the predictor has already learned the target-domain distribution, making \method\;a test-time specialization stage rather than only a mechanism for correcting domain shifts.

\subsection{\method \; Improves Spectral Purity by Suppressing Spurious Peaks}

We next examine how \method{} redistributes intensity within the frozen fragmentation space. For peak-quality evaluation on M+H and M+Na, we follow the Iceberg evaluation setting: retain the 100 highest-intensity predicted bins, remove peaks below 1\% of the predicted base-peak intensity, and match the remaining bins to the experimental spectrum. Both \method{} and \method--TTT improve intensity-weighted precision and F1 (Table~\ref{tab:ranking}), concentrating more predicted intensity on experimentally supported peaks. We further rank the saved M+Na top-100 predictions by intensity and assess agreement with experimental peak support and intensities using average precision (AP), NDCG, and Spearman correlation on matched bins. Improvements across these measures indicate better prioritization of supported peaks and more faithful relative intensity ordering.

Besides, we compare Iceberg and the validation-selected SPARC checkpoint over the complete candidate space, progressively adding peaks in descending predicted-intensity order until matched experimental peaks account for 25\%, 50\%, 75\%, or 100\% of the experimental ion current recoverable within the frozen candidate space. We compare purity at each coverage level and find that purity increases at full recoverable coverage, showing that less unsupported predicted intensity accompanies comparable experimental signal recovery. 
\begin{table}[!htbp]
\centering
\caption{Peak quality, candidate ranking, and spectral purity of Iceberg and \method{}.}
\label{tab:ranking}
\footnotesize
\setlength{\tabcolsep}{2.0pt}
\renewcommand{\arraystretch}{1.00}

\makebox[\columnwidth][c]{%
\begin{tabular}{@{}l|rrr|rrr|rrr|rrr@{}}
    \toprule
    & \multicolumn{3}{c|}{MSG M+H Peak}
    & \multicolumn{3}{c|}{MSG M+H Inten}
    & \multicolumn{3}{c|}{MSG M+Na Peak}
    & \multicolumn{3}{c}{MSG M+Na Inten} \\
    Method
    & Prec. & Recall & F1
    & Prec. & Recall & F1
    & Prec. & Recall & F1
    & Prec. & Recall & F1 \\
    \midrule

    Iceberg
    & 0.203 & \textbf{0.550} & 0.297
    & 0.465 & \textbf{0.756} & 0.576
    & 0.039 & \textbf{0.159} & 0.062
    & 0.153 & \textbf{0.314} & 0.206 \\

    \method
    & 0.264 & 0.449 & 0.333
    & 0.581 & 0.691 & 0.632
    & \textbf{0.061} & 0.097 & \textbf{0.075}
    & \textbf{0.361} & 0.297 & 0.326 \\

    \method--TTT
    & \textbf{0.271} & 0.438 & \textbf{0.335}
    & \textbf{0.593} & 0.687 & \textbf{0.637}
    & 0.045 & 0.112 & 0.074
    & 0.340 & 0.312 & \textbf{0.327} \\

    \bottomrule
\end{tabular}%
}

\vspace{3pt}

\makebox[\columnwidth][c]{%
\begin{tabular}{@{}l|rrrr|rrrr|rrrr@{}}
    \toprule
    & \multicolumn{4}{c|}{MSG M+Na Ranking}
    & \multicolumn{4}{c|}{MSG M+Na Purity}
    & \multicolumn{4}{c}{MSG M+H Purity} \\
    Method
    & AP & N@10 & N@50 & Spear
    & P@25 & P@50 & P@75 & P@100
    & P@25 & P@50 & P@75 & P@100 \\
    \midrule

    Iceberg
    & 0.402 & 0.508 & 0.567 & 0.268
    & 0.459 & 0.444 & 0.414 & 0.383
    & 0.804 & 0.765 & 0.722 & 0.664 \\

    \method
    & 0.474 & 0.612 & 0.648 & 0.409
    & \textbf{0.561} & \textbf{0.573}
    & \textbf{0.569} & \textbf{0.567}
    & \textbf{0.809} & \textbf{0.771}
    & \textbf{0.740} & \textbf{0.699} \\

    \method--TTT
    & \textbf{0.485} & \textbf{0.624}
    & \textbf{0.664} & \textbf{0.410}
    & -- & -- & -- & --
    & -- & -- & -- & -- \\

    \bottomrule
\end{tabular}%
}
\end{table}

\begin{figure}[!hbtp]
  \centering
  \includegraphics[width=\columnwidth]{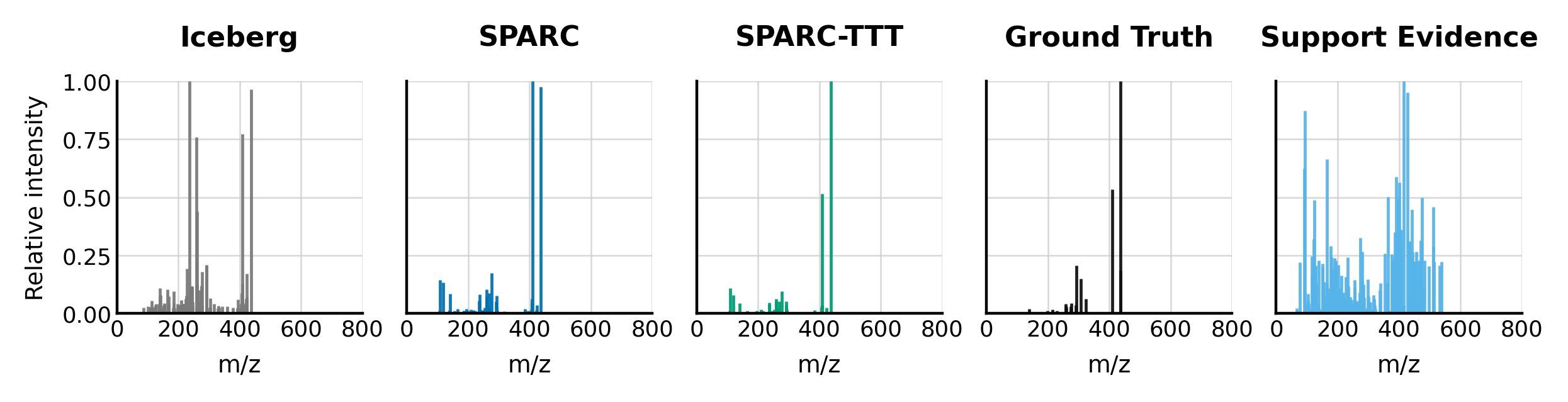}
  \caption{\textbf{Representative example of prediction from \method.} Each displayed spectrum is normalized, and peaks with intensities below 1\% are removed.}
  \label{fig:mechanism}
\end{figure}

Fig. ~\ref{fig:mechanism} illustrates this effect: the source
prediction assigns substantial intensity to low-$m/z$ peaks, whereas
the adapted predictions suppress many of these peaks and concentrate
relative intensity on the dominant peak. Together, these
results show that useful target-domain corrections can be made within
the existing candidate space by changing which fragments receive
substantial intensity.

\subsection{\method\ Transfers to Application-Specific Target Domains}

We next evaluate target-only specialization on five application-specific GNPS
libraries using MSG-training spectra as references, without a validation-specialization stage. Retrieval follows the acquisition-aware policy in Sec. \ref{sec:chemical-retrieval}, and online updates use entropy-based rollback.

Measured target-library spectra are reserved for evaluation. \method{} achieves the best EntSim among the compared predictors on every library and the best scores across all three reported metrics on GNPS-A/B and SC-FDA. These results demonstrate specialization to application-defined query streams using an external reference library.

\begin{table}[!hbtp]
\centering
\caption{Performance on Drugs of Abuse, 3HAA, ECG, GNPS-A/B, and SC-FDA. All models are initially trained on the MSG dataset.}
\label{tab:full_reference}
\scriptsize
\setlength{\tabcolsep}{2pt}
\renewcommand{\arraystretch}{0.9}
\resizebox{\columnwidth}{!}{%
\begin{tabular}{@{}l|rrr|rrr|rrr|rrr|rrr@{}}
    \toprule
    & \multicolumn{3}{|c}{DoA}
    & \multicolumn{3}{|c}{3HAA}
    & \multicolumn{3}{|c}{ECG}
    & \multicolumn{3}{|c}{GNPS-A/B}
    & \multicolumn{3}{|c}{SC-FDA} \\
    Method
    & MSE & EntSim & Cos
    & MSE & EntSim & Cos
    & MSE & EntSim & Cos
    & MSE & EntSim & Cos
    & MSE & EntSim & Cos \\
    \midrule
    NEIMS
    & 19.90 & 0.198 & 0.127
    & 12.84 & 0.386 & 0.382
    & 34.33 & 0.184 & 0.145
    & 18.05 & 0.409 & 0.242
    & 14.75 & 0.148 & 0.110 \\

    GrAFF-MS
    & 19.98 & 0.149 & 0.084
    & 15.23 & 0.238 & 0.173
    & 34.49 & 0.145 & 0.117
    & 19.84 & 0.324 & 0.179
    & 15.57 & 0.157 & 0.107 \\

    FIORA
    & 1.078 & 0.427 & 0.458
    & 0.935 & 0.395 & 0.397
    & 2.836 & 0.140 & 0.117
    & 1.009 & 0.393 & 0.488
    & \underline{0.893} & 0.362 & \underline{0.405} \\

    Iceberg
    & \textbf{1.057} & 0.415 & 0.483
    & \underline{0.904} & 0.490 & \underline{0.509}
    & \textbf{2.621} & 0.178 & 0.223
    & \underline{0.960} & 0.504 & 0.511
    & 0.925 & 0.322 & 0.379 \\

    \method
    & \underline{1.058} & \textbf{0.482} & \textbf{0.509}
    & \textbf{0.901} & \textbf{0.508} & \textbf{0.514}
    & 2.654 & \textbf{0.205} & \underline{0.246}
    & \textbf{0.919} & \textbf{0.534} & \textbf{0.569}
    & \textbf{0.839} & \textbf{0.389} & \textbf{0.435} \\
    
    \midrule

    w/o rollback
    & 1.068 & \underline{0.451} & \underline{0.488}
    & 0.940 & \underline{0.493} & 0.497
    & \underline{2.631} & \underline{0.202} & \textbf{0.252}
    & 1.109 & \underline{0.516} & \underline{0.536}
    & 0.920 & \underline{0.367} & 0.391 \\
    \bottomrule
\end{tabular}%
}
\end{table}

\subsection{Component Ablation}

Table~\ref{tab:component-ablation} compares component ablations under adaptation on the MSG $\rightarrow$ MSG-MNa transfer. All tested ablations reduce CosSim relative to the full method. Removing similarity-based support retrieval produces the largest decrease, whereas removing stochastic restoration has the smallest effect. These comparisons identify similarity-based support use as the most influential component among the tested ablations.

\WFclear
\begin{wraptable}{r}{0.6\columnwidth}
  \vspace{-1.2em}
  \centering
  \caption{\textbf{Ablation Study of \method\;Components.} }
  \label{tab:component-ablation}
  \small
  \setlength{\tabcolsep}{3pt}
  \resizebox{\linewidth}{!}{%
  \begin{tabular}{lrrr}
    \toprule
    Method & MSE & EntSim & CosSim \\
    \midrule
    -similarity-based support retrieval & 0.9859 & 0.2595 & 0.3441 \\
    -support-calibrated bin confidence & 0.8889 & 0.2637 & 0.3514 \\
    -stochastic restoration & 0.8840 & 0.2720 & 0.3692 \\
    -per-query rollback & 0.8977 & 0.2656 & 0.3631 \\
    \method & \textbf{0.8804} & \textbf{0.2742} & \textbf{0.3755} \\
    \bottomrule
  \end{tabular}%
  }
  \vspace{-1em}
\end{wraptable}

These results support the roles of chemically relevant references, direct support supervision, and stochastic restoration in continual adaptation. Removing similarity-based support produces the clearest degradation, indicating that adaptation depends largely on local chemical relevance. Removing stochastic restoration leads to the smallest decrease in CosSim, suggesting that this mechanism is not uniformly necessary and that allowing the adapted model to move farther from the source parameters may be beneficial for mass spectrum prediction.

\section{Limitations}


\method{} has two limitations. First, it is bounded by the fragmentation space inherited from the pretrained predictor. Freezing the fragment generator stabilizes adaptation but restricts \method\;to reweighting candidate fragments and it can't recover an omitted diagnostic fragment. Jointly expanding the candidate space without destabilizing online adaptation can be an important direction for domains with unseen fragmentation pathways.

Second, \method{} assumes that chemically similar, acquisition-compatible support spectra provide transferable supervision. Poor support coverage or a fingerprint-caused mismatch can bias both the supervised update and the support-derived reliability estimate. Continual updates further introduce dependence on target-stream order, while entropy-based rollback limits abrupt drift without guaranteeing correct peak locations. Query-specific uncertainty estimates and order-robust safeguards aligned more directly with spectral fidelity could mitigate these limitations.
\section{Conclusion}

We introduced SPARC, a retrieval-guided test-time specialization framework for
pretrained MS/MS predictors. \method{} preserves the fragmentation space and adapts fragment intensities using chemically related reference spectra, with support reconstruction errors weighting teacher consistency. Validation-assisted specialization improves benchmark spectral similarity, with further gains from online continuation. Without validation specialization, entropy-guarded adaptation using MSG references improves EntSim across all GNPS application libraries. Peak-level analyses associate these gains with higher spectral purity and intensity redistribution within the existing candidate space. Together, these results support reference-guided specialization as an extension of pretrained predictors when compatible labeled reference spectra are available.

\clearpage
\bibliography{iclr2027_conference}

@article{iceberg_repo,
  title   = {Generating Molecular Fragmentation Graphs with Autoregressive Neural Networks},
  author  = {Goldman, Samuel and Li, Janet and Coley, Connor W.},
  journal = {Analytical Chemistry},
  volume  = {96},
  number  = {8},
  pages   = {3419--3428},
  year    = {2024},
  doi     = {10.1021/acs.analchem.3c04654},
}

@article{young2024fragnnet,
  title={FraGNNet: a deep probabilistic model for tandem mass spectrum prediction},
  author={Young, Adamo and Wang, Fei and Wishart, David S and Wang, Bo and Greiner, Russell and R{\"o}st, Hannes},
  journal={arXiv preprint arXiv:2404.02360},
  year={2024}
}

@article{ye2024testtime,
  title   = {Test-Time Training for Deep {MS/MS} Spectrum Prediction Improves Peptide Identification},
  author  = {Ye, Jianbai and He, Xiangnan and Wang, Shujuan and Dong, Meng-Qiu and Wu, Feng and Lu, Shan and Feng, Fuli},
  journal = {Journal of Proteome Research},
  volume  = {23},
  number  = {2},
  pages   = {550--559},
  year    = {2024},
  doi     = {10.1021/acs.jproteome.3c00229},
  pmid    = {38153036}
}

@article{mismetti2026testtime,
  title         = {Test-Time Tuned Language Models Enable End-to-End De Novo Molecular Structure Generation from {MS/MS} Spectra},
  author        = {Mismetti, Laura and Alberts, Marvin and Krause, Andreas and Graziani, Mara},
  journal       = {arXiv preprint arXiv:2510.23746},
  year          = {2026},
  eprint        = {2510.23746},
  archiveprefix = {arXiv},
  primaryclass  = {cs.LG},
  url           = {https://arxiv.org/abs/2510.23746}
}

@article{bushuiev2024massspecgym,
  title={MassSpecGym: A benchmark for the discovery and identification of molecules},
  author={Bushuiev, Roman and Bushuiev, Anton and de Jonge, Niek F and Young, Adamo and Kretschmer, Fleming and Samusevich, Raman and Heirman, Janne and Wang, Fei and Zhang, Luke and D{\"u}hrkop, Kai and others},
  journal={Advances in Neural Information Processing Systems},
  volume={37},
  pages={110010--110027},
  year={2024}
}

@article{wang2021tent,
  title={Tent: Fully test-time adaptation by entropy minimization},
  author={Wang, Dequan and Shelhamer, Evan and Liu, Shaoteng and Olshausen, Bruno and Darrell, Trevor},
  journal={arXiv preprint arXiv:2006.10726},
  year={2020}
}

@inproceedings{wang2022cotta,
  title={Continual test-time domain adaptation},
  author={Wang, Qin and Fink, Olga and Van Gool, Luc and Dai, Dengxin},
  booktitle={2022 IEEE/CVF Conference on Computer Vision and Pattern Recognition (CVPR)},
  pages={7191--7201},
  year={2022},
  organization={IEEE}
}

@article{cui2025taip,
  title   = {Online Test-Time Adaptation for Better Generalization of Interatomic Potentials to Out-of-Distribution Data},
  author  = {Cui, Taoyong and Tang, Chenyu and Zhou, Dongzhan and Li, Yuqiang and Gong, Xingao and Ouyang, Wanli and Su, Mao and Zhang, Shufei},
  journal = {Nature Communications},
  volume  = {16},
  pages   = {1891},
  year    = {2025},
  doi     = {10.1038/s41467-025-57101-4}
}

@article{wang2016gnps,
  title   = {Sharing and Community Curation of Mass Spectrometry Data with Global Natural Products Social Molecular Networking},
  author  = {Wang, Mingxun and Carver, Jeremy J. and Phelan, Vanessa V. and others},
  journal = {Nature Biotechnology},
  volume  = {34},
  number  = {8},
  pages   = {828--837},
  year    = {2016},
  doi     = {10.1038/nbt.3597}
}

@article{horai2010massbank,
  title   = {{MassBank}: A Public Repository for Sharing Mass Spectral Data for Life Sciences},
  author  = {Horai, Hisayuki and Arita, Masanori and Kanaya, Shigehiko and others},
  journal = {Journal of Mass Spectrometry},
  volume  = {45},
  number  = {7},
  pages   = {703--714},
  year    = {2010},
  doi     = {10.1002/jms.1777}
}

@article{duhrkop2021canopus,
  title   = {Systematic Classification of Unknown Metabolites Using High-Resolution Fragmentation Mass Spectra},
  author  = {D{\"u}hrkop, Kai and Nothias, Louis-F{\'e}lix and Fleischauer, Markus and others},
  journal = {Nature Biotechnology},
  volume  = {39},
  number  = {4},
  pages   = {462--471},
  year    = {2021},
  doi     = {10.1038/s41587-020-0740-8}
}

@article{brungs2025msnlib,
  title   = {{$\mathrm{MS}^{n}$Lib}: Efficient Generation of Open Multi-Stage Fragmentation Mass Spectral Libraries},
  author  = {Brungs, Corinna and Schmid, Robin and Heuckeroth, Steffen and others},
  journal = {Nature Methods},
  volume  = {22},
  pages   = {2028--2031},
  year    = {2025},
  doi     = {10.1038/s41592-025-02813-0}
}

@article{gupta2026spectraverse,
  title   = {Comprehensive Curation and Harmonization of Small-Molecule {MS/MS} Libraries in {Spectraverse}},
  author  = {Gupta, Vishu and Qiang, Hantao and Chung, Hsin-Hsiang and Herbst, Ehud and Skinnider, Michael A.},
  journal = {Analytical Chemistry},
  volume  = {98},
  number  = {5},
  pages   = {3934--3943},
  year    = {2026},
  doi     = {10.1021/acs.analchem.5c06256}
}

@article{elloumi2024neomsms,
  title   = {From {MS/MS} Library Implementation to Molecular Networks: Exploring Oxylipin Diversity with {NEO-MSMS}},
  author  = {Elloumi, Anis and Mas-Normand, Lindsay and Bride, Jamie and others},
  journal = {Scientific Data},
  volume  = {11},
  pages   = {193},
  year    = {2024},
  doi     = {10.1038/s41597-024-03034-4}
}

@article{yurekten2024metabolights,
  title={MetaboLights: open data repository for metabolomics},
  author={Yurekten, Ozgur and Payne, Thomas and Tejera, Noemi and Amaladoss, Felix Xavier and Martin, Callum and Williams, Mark and O’Donovan, Claire},
  journal={Nucleic acids research},
  volume={52},
  number={D1},
  pages={D640--D646},
  year={2024},
  publisher={Oxford University Press}
}

@article{duhrkop2019sirius4,
  title   = {{SIRIUS} 4: A Rapid Tool for Turning Tandem Mass Spectra into Metabolite Structure Information},
  author  = {D{\"u}hrkop, Kai and Fleischauer, Markus and Ludwig, Marcus and Aksenov, Alexander A. and Melnik, Alexey V. and Meusel, Marvin and Dorrestein, Pieter C. and Rousu, Juho and B{\"o}cker, Sebastian},
  journal = {Nature Methods},
  volume  = {16},
  number  = {4},
  pages   = {299--302},
  year    = {2019},
  doi     = {10.1038/s41592-019-0344-8}
}

@article{schymanski2014confidence,
  title   = {Identifying Small Molecules via High Resolution Mass Spectrometry: Communicating Confidence},
  author  = {Schymanski, Emma L. and Jeon, Junho and Gulde, Rebekka and Fenner, Kathrin and Ruff, Matthias and Singer, Heinz P. and Hollender, Juliane},
  journal = {Environmental Science \& Technology},
  volume  = {48},
  number  = {4},
  pages   = {2097--2098},
  year    = {2014},
  doi     = {10.1021/es5002105}
}

@inproceedings{murphy2023graffms,
  title     = {Efficiently Predicting High Resolution Mass Spectra with Graph Neural Networks},
  author    = {Murphy, Michael and Jegelka, Stefanie and Fraenkel, Ernest and Kind, Tobias and Healey, David and Butler, Thomas},
  booktitle = {Proceedings of the 40th International Conference on Machine Learning},
  series    = {Proceedings of Machine Learning Research},
  volume    = {202},
  pages     = {25549--25562},
  publisher = {PMLR},
  year      = {2023}
}

@article{young2024massformer,
  title   = {Tandem Mass Spectrum Prediction for Small Molecules Using Graph Transformers},
  author  = {Young, Adamo and R{\"o}st, Hannes and Wang, Bo},
  journal = {Nature Machine Intelligence},
  volume  = {6},
  number  = {4},
  pages   = {404--416},
  year    = {2024},
  doi     = {10.1038/s42256-024-00816-8}
}

@article{bremer2022cfmid,
  title   = {How Well Can We Predict Mass Spectra from Structures? Benchmarking Competitive Fragmentation Modeling for Metabolite Identification on Untrained Tandem Mass Spectra},
  author  = {Bremer, Parker Ladd and Vaniya, Arpana and Kind, Tobias and Wang, Shunyang and Fiehn, Oliver},
  journal = {Journal of Chemical Information and Modeling},
  volume  = {62},
  number  = {17},
  pages   = {4049--4056},
  year    = {2022},
  doi     = {10.1021/acs.jcim.2c00936}
}

@article{hoang2024platforms,
  title   = {Tandem Mass Spectrometry across Platforms},
  author  = {Hoang, Corey and Uritboonthai, Winnie and Hoang, Linh and Billings, Elizabeth M. and Aisporna, Aries and Nia, Farshad A. and Derks, Rico J. E. and Williamson, James R. and Giera, Martin and Siuzdak, Gary},
  journal = {Analytical Chemistry},
  volume  = {96},
  number  = {14},
  pages   = {5478--5488},
  year    = {2024},
  doi     = {10.1021/acs.analchem.3c05576}
}

@article{li2021spectralentropy,
  title   = {Spectral Entropy Outperforms {MS/MS} Dot Product Similarity for Small-Molecule Compound Identification},
  author  = {Li, Yuanyue and Kind, Tobias and Folz, Jacob and Vaniya, Arpana and Mehta, Sajjan Singh and Fiehn, Oliver},
  journal = {Nature Methods},
  volume  = {18},
  number  = {12},
  pages   = {1524--1531},
  year    = {2021},
  doi     = {10.1038/s41592-021-01331-z}
}

@article{wei2019rapid,
  title={Rapid Prediction of Electron--Ionization Mass Spectrometry Using Neural Networks},
  author={Wei, Jennifer N and Belanger, David and Adams, Ryan P and Sculley, D},
  journal={ACS Central Science},
  volume={5},
  number={4},
  pages={700--708},
  year={2019},
  doi={10.1021/acscentsci.9b00085}
}

@article{hong2023dmolms,
  title={3DMolMS: prediction of tandem mass spectra from 3D molecular conformations},
  author={Hong, Yuhui and Li, Sujun and Welch, Christopher J and Tichy, Shane and Ye, Yuzhen and Tang, Haixu},
  journal={Bioinformatics},
  volume={39},
  number={6},
  pages={btad354},
  year={2023},
  doi={10.1093/bioinformatics/btad354}
}

@article{nowatzky2025fiora,
  title={FIORA: Local neighborhood-based prediction of compound mass spectra from single fragmentation events},
  author={Nowatzky, Yannek and Russo, Francesco Friedrich and Lisec, Jan and Kister, Alexander and Reinert, Knut and Muth, Thilo and Benner, Philipp},
  journal={Nature Communications},
  volume={16},
  pages={2298},
  year={2025},
  doi={10.1038/s41467-025-57422-4}
}

@article{bushuiev2026self,
  title={Self-supervised learning of molecular representations from millions of tandem mass spectra using DreaMS},
  author={Bushuiev, Roman and Bushuiev, Anton and Samusevich, Raman and Brungs, Corinna and Sivic, Josef and Pluskal, Tom{\'a}{\v{s}}},
  journal={Nature Biotechnology},
  volume={44},
  number={4},
  pages={630--640},
  year={2026},
  publisher={Nature Publishing Group US New York}
}

@article{liu2025adduct,
  title   = {Adduct-Induced Variability in Tandem Mass Spectrometry},
  author  = {Liu, Botao and Tang, Zhifeng and Huan, Tao},
  journal = {Analytical Chemistry},
  volume  = {97},
  number  = {31},
  pages   = {17058--17066},
  year    = {2025},
  doi     = {10.1021/acs.analchem.5c02792}
}

@article{schmid2021iimn,
  title   = {Ion Identity Molecular Networking for Mass Spectrometry-Based
             Metabolomics in the {GNPS} Environment},
  author  = {Schmid, Robin and Petras, Daniel and Nothias, Louis-Felix
             and Wang, Mingxun and Aron, Allegra T. and others},
  journal = {Nature Communications},
  volume  = {12},
  pages   = {3832},
  year    = {2021},
  doi     = {10.1038/s41467-021-23953-9}
}

@article{huber2021spec2vec,
  title   = {{Spec2Vec}: Improved Mass Spectral Similarity Scoring through
             Learning of Structural Relationships},
  author  = {Huber, Florian and Ridder, Lars and Verhoeven, Stefan
             and Spaaks, Jurriaan H. and Diblen, Faruk and Rogers, Simon
             and van der Hooft, Justin J. J.},
  journal = {PLOS Computational Biology},
  volume  = {17},
  number  = {2},
  pages   = {e1008724},
  year    = {2021},
  doi     = {10.1371/journal.pcbi.1008724}
}

@article{dejonge2023ms2query,
  title   = {{MS2Query}: Reliable and Scalable {MS2} Mass Spectra-Based
             Analogue Search},
  author  = {de Jonge, Niek F. and Louwen, Joris J. R. and Chekmeneva, Elena
             and Camuzeaux, Stephane and Vermeir, Femke J. and Jansen, Robert S.
             and Huber, Florian and van der Hooft, Justin J. J.},
  journal = {Nature Communications},
  volume  = {14},
  pages   = {1752},
  year    = {2023},
  doi     = {10.1038/s41467-023-37446-4}
}

@article{rogers2010ecfp,
  title   = {Extended-Connectivity Fingerprints},
  author  = {Rogers, David and Hahn, Mathew},
  journal = {Journal of Chemical Information and Modeling},
  volume  = {50},
  number  = {5},
  pages   = {742--754},
  year    = {2010},
  doi     = {10.1021/ci100050t}
}

@article{carhart1985atompairs,
  title   = {Atom Pairs as Molecular Features in Structure-Activity Studies:
             Definition and Applications},
  author  = {Carhart, Raymond E. and Smith, Dennis H. and
             Venkataraghavan, R.},
  journal = {Journal of Chemical Information and Computer Sciences},
  volume  = {25},
  number  = {2},
  pages   = {64--73},
  year    = {1985},
  doi     = {10.1021/ci00046a002}
}

@inproceedings{tarvainen2017meanteacher,
  title     = {Mean Teachers Are Better Role Models:
               Weight-Averaged Consistency Targets Improve
               Semi-Supervised Deep Learning Results},
  author    = {Tarvainen, Antti and Valpola, Harri},
  booktitle = {Advances in Neural Information Processing Systems},
  volume    = {30},
  year      = {2017}
}

@article{bemis1996murcko,
  title   = {The Properties of Known Drugs. 1. Molecular Frameworks},
  author  = {Bemis, Guy W. and Murcko, Mark A.},
  journal = {Journal of Medicinal Chemistry},
  volume  = {39},
  number  = {15},
  pages   = {2887--2893},
  year    = {1996},
  doi     = {10.1021/jm9602928}
}

@article{kruve2013sodium,
  title   = {Sodium Adduct Formation Efficiency in {ESI} Source},
  author  = {Kruve, Anneli and Kaupmees, Karl and Liigand, Jaanus
             and Oss, Merit and Leito, Ivo},
  journal = {Journal of Mass Spectrometry},
  volume  = {48},
  number  = {6},
  pages   = {695--702},
  year    = {2013},
  doi     = {10.1002/jms.3218}
}

@article{dejonge2026crossionization,
  title   = {Cross Ionization Mode Chemical Similarity Prediction between Tandem Mass Spectra in Metabolomics},
  author  = {de Jonge, Niek F. and Chekmeneva, Elena and Schmid, Robin
             and Joas, David and Truong, Lem-Joe and van der Hooft, Justin J. J.
             and Huber, Florian},
  journal = {Nature Communications},
  volume  = {17},
  pages   = {2483},
  year    = {2026}
}

@article{huber2021ms2deepscore,
  title   = {{MS2DeepScore}: A Novel Deep Learning Similarity Measure to Compare Tandem Mass Spectra},
  author  = {Huber, Florian and van der Burg, Sven and van der Hooft, Justin J. J. and Ridder, Lars},
  journal = {Journal of Cheminformatics},
  volume  = {13},
  pages   = {84},
  year    = {2021},
  doi     = {10.1186/s13321-021-00558-4}
}
\bibliographystyle{iclr2027_conference}

\end{document}